%% file: main.tex
\documentclass{article}

\PassOptionsToPackage{numbers,sort&compress}{natbib}
\usepackage[preprint]{neurips_2026}

\usepackage[utf8]{inputenc}
\usepackage[T1]{fontenc}
\usepackage{hyperref}
\usepackage{url}
\usepackage{booktabs}
\usepackage{amsmath}
\usepackage{amssymb}
\usepackage{graphicx}
\usepackage{xcolor}
\usepackage{microtype}
\usepackage{nicefrac}
\usepackage{tikz}
\usepackage{pgfplots}
\usepackage{subcaption}
\usepackage[capitalize]{cleveref}
\usepackage{xcolor}

\usepackage{wrapfig}
\usepackage{placeins}
\usepackage{enumitem}
\setlist[itemize]{leftmargin=*, itemsep=2pt, parsep=0pt, topsep=2pt}
\setlist[enumerate]{leftmargin=*, itemsep=2pt, parsep=0pt, topsep=2pt}

\pgfplotsset{compat=1.18}
\usetikzlibrary{arrows.meta,positioning,decorations.pathreplacing,calc,backgrounds,fit}
\graphicspath{{figures/assets/}}

\crefname{section}{Sec.}{Secs.}
\Crefname{section}{Section}{Sections}
\Crefname{table}{Table}{Tables}
\crefname{table}{Tab.}{Tabs.}

\title{Transcoda: End-to-End Zero-Shot Optical Music Recognition via Data-Centric Synthetic Training}

\author{%
  \normalfont
  Daniel Dratschuk \; and \; Paul Swoboda \\[4pt]
  Heinrich Heine University Düsseldorf, Germany \\
  \texttt{\{daniel.dratschuk,paul.swoboda\}@hhu.de} \\
}

\begin{document}

\maketitle

\begin{abstract}
    Optical Music Recognition (OMR), the task of transcribing sheet music into a structured textual representation, is currently bottlenecked by a lack of large-scale, annotated datasets of real scans. This forces models to rely on either few-shot transfer or synthetic training pipelines that remain overly simplistic. A secondary challenge is encoding non-uniqueness: in the popular Humdrum \texttt{**kern} format for transcribing music, multiple different text encodings can render into the same visual sheet music. This one-to-many mapping creates a harder learning task and introduces high uncertainty during decoding. We propose Transcoda, an OMR system built on (i) an advanced synthetic data generation pipeline, (ii) a normalization of the \texttt{**kern} encoding to enforce a unique normal form and (iii) grammar-based decoding to ensure the syntactic correctness of the output. This approach allows us to train a compact 59M-parameter model in just 6 hours on a single GPU that outperforms billion-parameter baselines. Transcoda achieves the best score among state of the art baselines on a newly curated benchmark of synthetically rendered scores at 18.46\% OMR-NED (compared to 43.91\% for the next-best system, Legato) and reduces the error rate on historical Polish scans to 63.97\% OMR-NED (down from 80.16\% for SMT++).
\end{abstract}

\input{figures/teaser}

\section{Introduction}
\label{sec:intro}

Optical Music Recognition (OMR) translates images of sheet music into
machine-readable textual formats. It is more complex than plain text
recognition because music notation is a dense two-dimensional graph: noteheads,
stems, beams, accidentals, staves, and rhythmic constraints interact across
space and time. Still, like for OCR, modern end-to-end systems treat OMR as
image-to-sequence generation, but unlike OCR they face a severe data problem.
Annotated datasets of physical scans are not publicly available at the scale
needed for current neural models, so systems are trained on synthetic
renderings and evaluated zero-shot on real scans
\cite{Yang2025Legato,RiosVila2026}.

For synthetic data the main failure mode is not only visual noise. Real scores
also have denser layouts and much longer textual targets than common synthetic
training samples. Due to the interdependency of different music voices, rhythm
etc.\ a small local visual error can push an autoregressive decoder into an
invalid continuation, after which the whole transcription can become unusable.

An additional challenge is non-uniqueness: The same visual score can map to
many syntactically different but semantically equivalent transcription
sequences. Flexible encoding conventions create a one-to-many training signal.
The decoder must model source-specific formatting variation in addition to the
musical structure. This makes training and decoding harder and distribution
shift additionally compounds uncertainty.

We introduce \textbf{Transcoda}, a compact 59M-parameter vision-encoder-decoder
model trained only on synthetic data encoded in the
\texttt{**kern}~\cite{Huron1997} format. Its core design choice is
data-centric. Before training, a deterministic pipeline normalizes textual
targets so that each rendered score maps to a unique grammar-aligned sequence.
At inference time, an optional constrained decoder can enforce formal
\texttt{**kern} validity whenever downstream rendering or playback needs it.
Our experiments show that normalization, not model scale or decoding
constraints, is the main stabilizer.

Our contributions are:
\begin{enumerate}
    \item We propose an elaborate synthetic data generation pipeline that better
      approaches the visual and structural complexity of real sheet music. We
      generate, to our knowledge, the largest openly available dataset for training
      OMR.
    \item We identify non-uniqueness as a bottleneck in zero-shot OMR and show that
      deterministic target normalization stabilizes autoregressive generation.
    \item We release a standardized synthetic evaluation protocol based on the Verovio
      rendering engine, and report ablations that disentangle sequence-modeling
      stability, visual domain shift, and formal validity.
    \item Empirically our compact end-to-end Transcoda model achieves the best score
      among state of the art publicly available OMR systems on our synthetic
      benchmark and improves over larger baselines on real historical scans.
\end{enumerate}

We release code, training scripts, and model weights at
\url{https://github.com/btrkeks/transcoda}.

\section{Related work}
\label{sec:related-work}

\paragraph{Traditional and modular OMR.}
Classical OMR pipelines decomposed the task into preprocessing, symbol detection, notation assembly, and encoding \cite{Rebelo2012}. These hand-crafted stages were fragile on handwritten or degraded documents \cite{Torras2021}, with errors cascading across stages \cite{Edirisooriya2021}. Deep learning was initially adopted to strengthen individual stages rather than to replace the pipeline itself: Pacha and Eidenberger trained a universal symbol classifier across aggregated datasets \cite{Pacha2017}, and Yang et al.\ jointly trained a notation-assembly model on imperfect YOLOv8 detector outputs to mitigate cascade effects \cite{Yang2024}. These systems improved component accuracy but inherited the staged structure and its failure modes.


\paragraph{Shift to End-to-End Architectures.}
To avoid cascading errors, the field shifted towards end-to-end systems that
treat OMR as a sequence recognition problem. Calvo-Zaragoza et al. proposed a
convolutional RNN that directly transcribes preprocessed
single-staff images into sequences of musical symbols \cite{Zaragoza2019}.
The adoption of Transformer architectures improved spatial modeling; Li et al.
proposed TrOMR to enhance contextual perception in polyphonic scores
\cite{TrOMR2023}, while Ríos-Vila et al. introduced the Sheet Music Transformer
(SMT) \cite{RiosVila2024SMT}, which was the first system to successfully transcribe entire single systems of pianoform scores without any simplifications

\paragraph{Recent Full-Page Baselines: SMT++ and Legato.}
Recent efforts have scaled end-to-end OMR from isolated systems to full pages
and beyond, establishing our two primary baselines. By implementing curriculum
learning, the SMT architecture was adapted into SMT++ \cite{RiosVila2026},
achieving the first full-page, end-to-end OMR for pianoform sheet music while
bypassing prior layout analysis or staff cropping. However, SMT++ relies on a
relatively small dataset, is computationally expensive to train, and frequently
generates syntactically invalid \texttt{**kern} notation. Expanding scale
further, Yang et al. introduced Legato \cite{Yang2025Legato}, which processes
multiple concatenated pages at a time. Legato uses a frozen Llama 3.2 11B
Vision encoder \cite{MetaLlama32Vision2024} and a custom Byte-Pair Encoding (BPE) tokenizer. Trained on over 214,000 synthetic scores, Legato directly generates
ABC notation \cite{Yang2025Legato,Walshaw2011ABC}.
Like \texttt{**kern}, ABC is a discrete text-based format for music, but it functions as a denser, more standardized alphanumeric shorthand.
Compared to SMT++, we use a simpler training protocol and substantially more training data; compared to Legato, a much smaller model with increased training data. Our data generation pipeline improves over both.
Nevertheless, our training and inference speed is still significantly higher.

\section{Method}
\label{sec:method}

Transcoda is a vision-encoder-decoder model that maps a fixed-size score image
to a \texttt{**kern} token sequence. The method has three parts: a compact
architecture, a target-normalization data engine, and an optional constrained
decoder.

\subsection{Architecture}
\label{sec:architecture}

\input{figures/architecture}

Our model processes fixed-resolution inputs of $1485 \times 1050$ pixels and
generates sequences up to a maximum target length of 2048 tokens. The network
contains approximately 58.8M total parameters, distributed across three core
components:
\begin{itemize}
    \item \textbf{Visual encoder (27.9M parameters):} We use a pretrained \texttt{facebook/convnextv2-tiny-22k-224} backbone \cite{Liu2022ConvNeXt}. We append 2D sinusoidal positional encodings \cite{Coquenet2023,SinghKarayev2021} to the output visual grid before flattening it into a sequence.
    \item \textbf{Projection bridge (2.6M parameters):} A two-layer MLP projects the flattened encoder features to match the decoder embedding dimension.
    \item \textbf{Autoregressive decoder (28.3M parameters):} We employ an 8-layer Pre-LN Transformer with $d_{\mathrm{model}}=512$, $d_{\mathrm{ff}}=1024$, and 8 attention heads. The decoder utilizes GELU feed-forward blocks and rotary positional embeddings (RoPE) for self-attention.
\end{itemize}

\textbf{In-domain visual pretraining.} To probe the visual domain gap, we add an optional unsupervised pretraining stage on $\sim$200,000 unlabelled historical score images from IMSLP \cite{ProjectPetrucci2025}, using a Fully Convolutional Masked Autoencoder (FCMAE) \cite{FCMAE}. Unlike the official sparse-convolution implementation, we adopt a dense SimMIM-style variant \cite{SimMIM}: learned mask tokens are injected after the patch embedding and the full dense ConvNeXt-V2 encoder runs over the masked features. This supports our fixed rectangular full-page canvas and lets us bias masking toward ink-heavy notation regions.

\subsection{\texttt{**kern} Tokenization} \label{sec:kern-background}

\textbf{\texttt{**kern}.} We adopt the Humdrum \texttt{**kern} format \cite{Huron1997} for encoding music.
\texttt{**kern} provides a dense, human-readable 2D matrix representation that minimizes token overhead while preserving complex polyphonic structures.
See Figure~\ref{fig:kern_format} for an illustration of the encoding.
Previous work, such as SMT++ \cite{RiosVila2026}, demonstrated its initial viability for OMR.

\input{figures/kern_format}
We chose \texttt{**kern} over the ubiquitous MusicXML format, since the latter relies on heavy XML boilerplate, which expands sequence lengths and drastically degrades autoregressive decoding efficiency.

\textbf{Compositional tokenization.}
We train a 3000-token Byte Pair Encoding (BPE) tokenizer over our normalized \texttt{**kern} dataset, while enforcing a strict split-space constraint during vocabulary construction.
Standard BPE would merge across spaces and memorize fixed representations of entire multi-note chords. Preventing cross-space merges forces a compositional representation, so the decoder can assemble novel chord combinations instead of failing on out-of-vocabulary vertical structures.

\subsection{Data Pipeline}
\label{sec:data-engine}

The data engine converts open textual corpora into \texttt{**kern}, filters invalid files, normalizes targets, renders pages and applies controlled
augmentations to explicitly bridge the visual sim-to-real gap.

\textbf{Stage 1: Format conversion.}
Large-scale symbolic music datasets, such as PDMX, are predominantly distributed
in MusicXML. Because Transcoda predicts Humdrum \texttt{**kern} sequences, we
convert these sources to our target representation using a patched fork of the
\texttt{musicxml2hum} reference utility\footnote{The standard implementation
    exhibited memory safety issues, causing segmentation faults on $>10\%$ of the
    PDMX corpus. We patched these parsing errors, reducing the failure rate to
    $<0.07\%$, and open-source our fork.}.

\input{figures/normalization}
\textbf{Stage 2: Filtering.}
We discard files with broken UTF-8, missing spine terminators, missing clefs,
severe conversion artifacts, impossible accidental runs, corrupted octave
spellings, or invalid measure mathematics.

\textbf{Stage 3: Target normalization.}
The \texttt{**kern} format allows for different encodings to render into the same music sheet.
To eliminate this non-uniqueness, we process each encoding with a 21-stage normalization pipeline enforcing a normalized form,
that ensures a near-deterministic mapping between visual input and textual encoding, see~\cref{tab:canonicalization_examples}.
We group the normalization passes into four primary operations:

\textit{Extraneous spine removal.}
We strip non-notation parallel spines (e.g., lyrics, dynamics) to ensure the
model strictly predicts core musical geometry only. To train robustness, we later
procedurally re-inject these elements during rendering as controllable visual
distractors.

\textit{Visual-semantic alignment.}
We systematically remove non-visual elements, such as playback-only grace rests
and terminal string markers. This guarantees that sequence length and token
content correlate directly with the rendered visual notation.

\textit{Syntactic token sorting.}
A single token often encodes multiple properties (duration, pitch, accidentals,
articulations) in arbitrary orders across different datasets. We enforce a
fixed, character-level sorting hierarchy for every token component and
explicitly sort chord notes in ascending pitch order.

\textit{Structural repair.}
We resolve contradictory annotations generated by upstream dataset converters.
We collapse conflicting accidentals (e.g., a sharp and a natural on the same
token) into single valid modifiers.

\begin{table}[h]
    \centering
    \caption{Examples of deterministic target normalization. Raw sequences with identical visual meanings are collapsed into a single form to reduce non-uniqueness.}
    \label{tab:canonicalization_examples}
    \resizebox{\columnwidth}{!}{%
        \begin{tabular}{@{}llll@{}}
            \toprule
            \textbf{Artifact Type} & \textbf{Raw Input}                            & \textbf{Normalized Output} & \textbf{Pipeline Action}          \\
            \midrule
            Syntactic Ambiguity    & \texttt{[8fJ} \textit{or} \texttt{8fJ[}       & \texttt{8f[J}             & Token syntax sorting              \\
            Contradictory XML      & \texttt{4c\#n}                                & \texttt{4cn}              & Resolve conflicting accidentals   \\
            Redundant Metadata     & \texttt{*met(c)} \textit{then} \texttt{*M4/4} & \texttt{*M4/4}            & Remove equivalent time signatures \\
            Self-Canceling         & \texttt{4G[]}                                 & \texttt{4G}               & Remove zero-length ties           \\
            \bottomrule
        \end{tabular}%
    }
\end{table}

\textbf{Stage 4: Rendering and augmentation.}
In order to make the visual appearance of our synthetically generated music
sheets more realistic, we heavily use augmentations.

\textit{Non-target notation injection.} We render expected performance markings, such as dynamics and tempo text, into the image without altering the underlying \texttt{**kern} sequence (\cref{fig:semantic_augmentation}). This trains the model to selectively transcribe the structural notes while ignoring natural but task-irrelevant musical symbols.

\textit{Raster-stage degradation.} We process the rendered images through a multi-stage offline pipeline to simulate physical scanning and aging artifacts. First, we apply geometric distortions—including affine translations, perspective warping, and spatial stretching using OpenCV \cite{opencv_library}, explicitly discarding transformations that push notation out of bounds. Second, we composite the transformed foreground onto synthetically generated paper backgrounds, introducing sampled textures, lighting gradients, and color casts. Finally, we apply extensive document-level degradations using Augraphy \cite{augraphy_paper,augraphy_library}. This stage injects physical print artifacts (e.g., ink bleed, mottling) and scanner noise (e.g., dirty rollers, shadow casting, and JPEG compression).
\input{figures/semantic_augmentation}

\subsection{Constrained decoding}
\label{sec:grammar-inference}

Some downstream tools require formally valid \texttt{**kern}, for example
rendering, which might break down when presented with small syntactical errors.
For this case, Transcoda includes an optional constrained decoder. A GBNF
grammar compiled with \texttt{xgrammar} masks locally invalid tokens
\cite{Dong2024XGrammar}. A Python-side logits processor then tracks global state that a local grammar cannot express: it dynamically maintains the active spine count and enforces consistent line width, masking tabs, newlines, and spine split/merge tokens that would violate it. At each step, invalid
continuations receive $-\infty$ logit mass.
The constraint stack currently uses greedy decoding. It guarantees formal validity, but can hurt raw edit distance when the visual alignment is wrong.

\section{Experiments}
\label{sec:experiments}

\subsection{Training}

We train with PyTorch Lightning and bfloat16 precision.
A train run from scratch on the
310,554 training examples and 7,583 held-out synthetic test examples takes 6 hours on one NVIDIA RTX 5090 GPU with 32 GB memory.
Optimization uses AdamW with effective batch size 72 and $(\beta_1,\beta_2) = (0.9, 0.999)$. The encoder learning rate is $3 \times 10^{-4}$, while the projector and decoder use $1 \times 10^{-3}$.
We use 500 warmup steps, cosine decay to $5 \times 10^{-5}$, weight decay 0.01 on weight matrices, gradient clipping at 1.0, and label smoothing 0.1.

\subsection{Datasets}

\input{figures/dataset_yield}
\textbf{Training Corpus.} To construct a diverse and large-scale training set, we compile raw score data from multiple open-source repositories to generate 310,554 examples. To increase score density and provide longer ground-truth transcription targets, we stochastically concatenate multiple short samples. We explicitly balance this concatenation to yield pages containing between one and six systems, subject to a maximum context limit of 2048 tokens. The source corpora consist of:

\textit{PDMX} \cite{long2024pdmx, xu2024generating}: A large-scale public domain dataset featuring high variance in score complexity.

\textit{Grandstaff} \cite{RiosVila2026}: A corpus of 41,598 pianoform scores.

\textit{MuseTrainer} \cite{musetrainer}: A curated library comprising 69 complex piano scores.

\textit{OpenScore Lieder \& Quartets} \cite{OpenScoreLieder, OpenScoreQuartet}: Collections of 19th-century vocal works (1,389 scores) and string quartets (122 scores).

\textbf{Evaluation Domains.} We evaluate model performance across three dataset splits to measure optimization stability, synthetic generalization, and robustness to real-world domain shifts:

\textit{Validation Set}: Used for model selection, this split consists of the Grandstaff validation partition, a 2\% held-out partition of the PDMX dataset, and the complete Polish historical dataset reference. Synthetic examples in this split are generated identically to the training data, but no visual augmentations are applied during validation.

\textit{Real Target Domain (Zero-Shot)}: To evaluate transfer to real scanned media, we report final metrics on a historical split of 102 Polish scanned scores \cite{PRAIGPolishScores}.

\textbf{Synthetic Rendering Details.}
We render the normalized \texttt{**kern} sequences into images using the Verovio Python library \cite{Pugin_2014,verovio_pypi}, scaling all outputs to a uniform resolution of $1485 \times 1050$ pixels.
To ensure that the model learns robust visual representations we heavily randomize the synthetic rendering parameters for each training example. Variables including font family, document scale, margins, line width, and system spacing are sampled uniformly. The full distribution of rendering ranges is detailed in \cref{app:rendering_settings}.

\subsection{Baselines}
We compare against SMT++ and Legato, the two currently best open source OMR systems.
To ensure a fair comparison, we evaluate the official public checkpoints without further fine-tuning, following their released decoding hyperparameters and preprocessing configurations.

    \paragraph{\textbf{SMT++}~\cite{RiosVila2026} (\url{https://huggingface.co/PRAIG/smt-fp-grandstaff})}
        SMT++ directly predicts Humdrum \texttt{**kern} notation. We restore the model's raw tokenized output (including special tokens for newlines and tabs) to valid \texttt{**kern} text. We do not convert this output to MusicXML; instead, we compute the OMR-NED metric directly on the \texttt{**kern} strings using \texttt{musicdiff} and \texttt{converter21}.

    \paragraph{\textbf{Legato}~\cite{Yang2025Legato} (\url{https://huggingface.co/guangyangmusic/legato})}
        While derived from MusicXML data, Legato is trained to output ABC notation. To compute the OMR-NED metric, we convert the predicted ABC strings to MusicXML using the \texttt{abc2xml.py} script and compare them against the ground-truth MusicXML.

\subsection{Inference and Decoding}
To ensure reproducibility, we explicitly define the generation hyperparameters for all evaluated models. We evaluate baselines at their officially recommended optimums.
\begin{itemize}
    \item \textbf{Transcoda (Ours):} We generate sequences up to a maximum length of 2048 using standard beam search with a width of 3. The optional constrained decoding engine is evaluated separately, as its purpose is enforcing formal validity rather than raw edit-distance optimization.
    \item \textbf{Legato:} Following the authors' official configuration, we decode ABC predictions using a beam width of 3, a repetition penalty of 1.1, and a maximum length of 2048.
    \item \textbf{SMT++:} We evaluate the raw tokenized \texttt{**kern} output using greedy decoding up to a maximum generation length of 2048.
\end{itemize}

\subsection{Metrics}

Because raw character-level metrics often fail to capture the spatial and semantic relationships inherent in musical notation, we evaluate our system using two specialized domain metrics alongside standard sequence comparison. Lower scores indicate better performance for all metrics.

\textbf{OMR Normalized Edit Distance (OMR-NED):} A format-agnostic metric that balances computational efficiency with perceptual meaningfulness \cite{Martinez2025}. Instead of comparing raw text tokens, OMR-NED computes the set edit distance between constituent musical symbols. It enforces strict temporal offset matching: notes, rests, and non-note directions are only directly compared if they occur at the exact same temporal position within a measure. Unmatched symbols are penalized as independent insertions and deletions.

\textbf{Tree Edit Distance with Note Flattening (TEDn):} Evaluates the hierarchical structure of predicted MusicXML files and correlates strongly with human evaluation \cite{HajicJr2016}. Standard tree edit distance disproportionately penalizes single-note errors because a single musical note contains many nested XML child nodes. TEDn mitigates this over-penalization by flattening note sub-trees into compact string representations (encoding pitch, duration, and stem direction) before computing the normalized edit distance \cite{ZhangShasha1989}.

\textbf{Character Error Rate (CER):} Measures the sequence-level Levenshtein distance between the predicted text and the ground truth. While computationally cheap, CER simplifies the score to a 1D text sequence and frequently fails to capture the magnitude of structural or hierarchical discrepancies. Furthermore, CER is highly sensitive to representational variance, as the same musical score can often be encoded using distinct, equally valid text sequences. We mitigate this issue through our strict data normalization.

\subsection{Results}

Quantitative results are shown in \Cref{fig:teaser}.
For an exemplary qualitative result see Figure~\ref{fig:qualitative}.

\textbf{In-domain synthetic performance.}
On the clean, standardized Verovio evaluation split, Transcoda significantly outperforms existing baselines (\cref{fig:teaser}).
Using beam search, Transcoda achieves an OMR-NED of 18.46\%, effectively halving the error rate of the heavily scaled Legato model (43.91\%) and drastically outperforming SMT++ (92.23\%).

\input{figures/ablation_results}

\textbf{Zero-shot transfer to real scans.}
All models suffer performance degradation under distribution shift to physical media. However, Transcoda remains the most robust. On historical Polish scans, our base model achieves a 63.97\% OMR-NED (beam search), compared to 80.16\% for SMT++ and 86.73\% for Legato.

\textbf{The impact of target normalization.}
\Cref{tab:ablations_indomain} isolates sequence modeling performance on clean synthetic data. The most critical finding is the impact of non-uniqueness: removing target normalization causes catastrophic sequence collapse, raising the OMR-NED from 18.71\% to 82.51\%. This confirms our core hypothesis that deterministic textual targets are strictly necessary to stabilize autoregressive generation in complex 2D OMR tasks.

\textbf{Data engine and length extrapolation.}
The ablation study on real scans (\cref{tab:ablations_transfer}) demonstrates the necessity of bridging both visual and structural gaps. Removing visual raster degradation and asymmetric semantic augmentations causes the OMR-NED to spike by 11.19 and 13.23 points, respectively. However, the most severe structural failure occurs when we remove score concatenation (+14.34 OMR-NED). Real physical scores are significantly longer and denser than standard synthetic samples. Without concatenating synthetic scores during training, the autoregressive decoder suffers from severe length extrapolation failures on physical pages. This proves that matching target density is just as critical as simulating visual noise.

\textbf{In-domain visual pretraining.}
To test the potential of closing the visual domain gap via unsupervised learning, we evaluated a larger ConvNeXt-V2-Base encoder pre-trained via our custom dense FCMAE setup. Despite being limited by compute to just two pretraining epochs, this setup yields a promising improvement in zero-shot transfer (63.97\% $\to$ 61.11\% OMR-NED). This provides a strong signal that scaled, domain-specific visual pretraining on raw archival data is a highly viable path for future work to further reduce the sim-to-real gap.

\textbf{Inference strategies.}
As shown in \cref{tab:ablations_indomain}, beam search slightly improves both CER (4.38 $\to$ 2.72) and OMR-NED (18.71 $\to$ 18.46) over greedy decoding. Constrained decoding yields negligible changes to the raw metrics but remains a crucial optional layer to guarantee formal structural validity for strict downstream rendering parsers.

\textbf{Qualitative evaluation.}
\Cref{fig:qualitative} illustrates specific failure modes on a physical scan. SMT++ fails early in the sequence: it predicts the wrong bottom clef and meter, which causes a cascade of pitch errors. Legato captures the broader structure but fails on fine syntactical details; it misclassifies a natural sign as a sharp, outputs incorrect rest durations and merges all beam groupings.
Transcoda produces a highly accurate transcription but exhibits a distinct structural bias: it omits courtesy accidentals. Because our normalized training data eliminates semantically redundant modifiers, the model correctly infers the underlying pitch but ignores the visually present natural sign. Furthermore, Transcoda begins to capture the correct internal beam subdivisions that the baselines completely ignore.

\input{figures/qualitative}

\section{Discussion and Limitations}
\label{sec:discussion}

Transcoda significantly improves zero-shot OMR performance over current public baselines.
The 63.97\% OMR-NED on historical Polish scans indicates additional room for improvement on real-world manuscript transcription.
Our granular evaluation reveals three distinct limitations:

\begin{enumerate}
    \item \textbf{Structural decoding failures:} The dominant source of error on real scans involves structural matrix misalignment. In dense pianoform textures, predictions frequently hallucinate Humdrum voice splits (\texttt{*{\textasciicircum}}) and merges (\texttt{*v}). Once a voice is improperly split, the autoregressive decoder drifts into incorrect line widths and loses horizontal musical alignment, failing to recover for the remainder of the page.
    \item \textbf{Rare synthetic runaway loops:} While aggregate synthetic performance is excellent, we observe a rare (approx.\ 1.18\%) failure mode where the model enters catastrophic generation loops. In these cases, the prediction length can exceed the target length by a factor of four, repeatedly outputting the same valid but hallucinated multi-line patterns. Because these loops consist of syntactically valid \texttt{**kern}, grammar-constrained decoding alone cannot prune them, suggesting the need for adaptive repetition penalties.
    \item \textbf{The semantic visual gap:} The worst-performing real scans feature extreme manuscript density, physical bleed-through, handwritten annotations, and complex chordal piano textures.
    We argue that either even more involved raster-degradations or a shift to pre-training on heavily degraded real scans from historical archives might further push performance.
    We refer to Appendix~\ref{app:polish_examples} for a few examples.
\end{enumerate}

\section{Conclusion}
\label{sec:conclusion}

We presented Transcoda, a compact end-to-end OMR model trained only on synthetic data that outperforms existing state of the art models with much fewer parameters and a simpler training protocol.
We have shown that the dominant lever for improved OMR results is better training data, including proper format (normalization), visual fidelity and semantic variety, not so much model scale or an involved training protocol.
We argue that for further improvements even better data generation might be the way to go.
We hope that Transcoda will advance technical possibilities in musicology, where potent open-source OMR tools are still lacking.

\FloatBarrier
{\small
    \bibliographystyle{plainnat}
    \bibliography{references}
}

\newpage
\appendix
\section{Synthetic Rendering Settings}
\label{app:rendering_settings}

Training images are rendered with Verovio 6.0.1. For each example, we sample
the rendering options in \Cref{tab:verovio_sampling}.

\begin{table}[h]
\centering
\small
\caption{Sampled Verovio rendering options used for synthetic training images.}
\label{tab:verovio_sampling}
\begin{tabular}{@{}ll@{}}
\toprule
Option & Sampling rule \\
\midrule
\texttt{scale} & Integer in $[52, 84]$ \\
\texttt{pageWidth} & $\texttt{image\_width} \cdot U(1.80, 2.40)$, with $\texttt{image\_width}=1050$ \\
\texttt{barLineWidth} & $U(0.16, 0.72)$ \\
\texttt{beamMaxSlope} & Integer in $[4, 16]$ \\
\texttt{staffLineWidth} & $U(0.10, 0.26)$ \\
\texttt{stemWidth} & $U(0.10, 0.36)$ \\
\texttt{ledgerLineThickness} & $U(0.10, 0.40)$ \\
\texttt{thickBarlineThickness} & $U(0.60, 1.80)$ \\
\texttt{spacingLinear} & $U(0.16, 0.32)$ \\
\texttt{spacingNonLinear} & $U(0.24, 0.56)$ \\
\texttt{spacingStaff} & Integer in $[4, 20]$ \\
\texttt{spacingSystem} & Integer in $[3, 10]$ \\
\texttt{measureMinWidth} & Integer in $[6, 18]$ \\
Page margins & $\texttt{pageWidth} \cdot U(0.015, 0.05)$, clamped to $[0, 500]$ \\
Font & Leipzig, Bravura, Gootville, Leland, or Petaluma \\
\bottomrule
\end{tabular}
\end{table}

We set \texttt{breaks=auto} and \texttt{footer=none}. The flags
\texttt{breaksNoWidow}, \texttt{justifyVertically}, and \texttt{noJustification}
are disabled. If rendering fails or produces an invalid layout, the retry path
keeps the same base recipe but tightens the page: it reduces scale, spacing,
and \texttt{measureMinWidth}, may increase \texttt{pageWidth}, and may reduce
margins.
\clearpage
\section{Polish Scan Examples}
\label{app:polish_examples}

\begin{figure}[!h]
    \centering
    \begin{subfigure}[t]{0.495\textwidth}
        \centering
        \includegraphics[width=\linewidth,height=0.36\textheight,keepaspectratio]{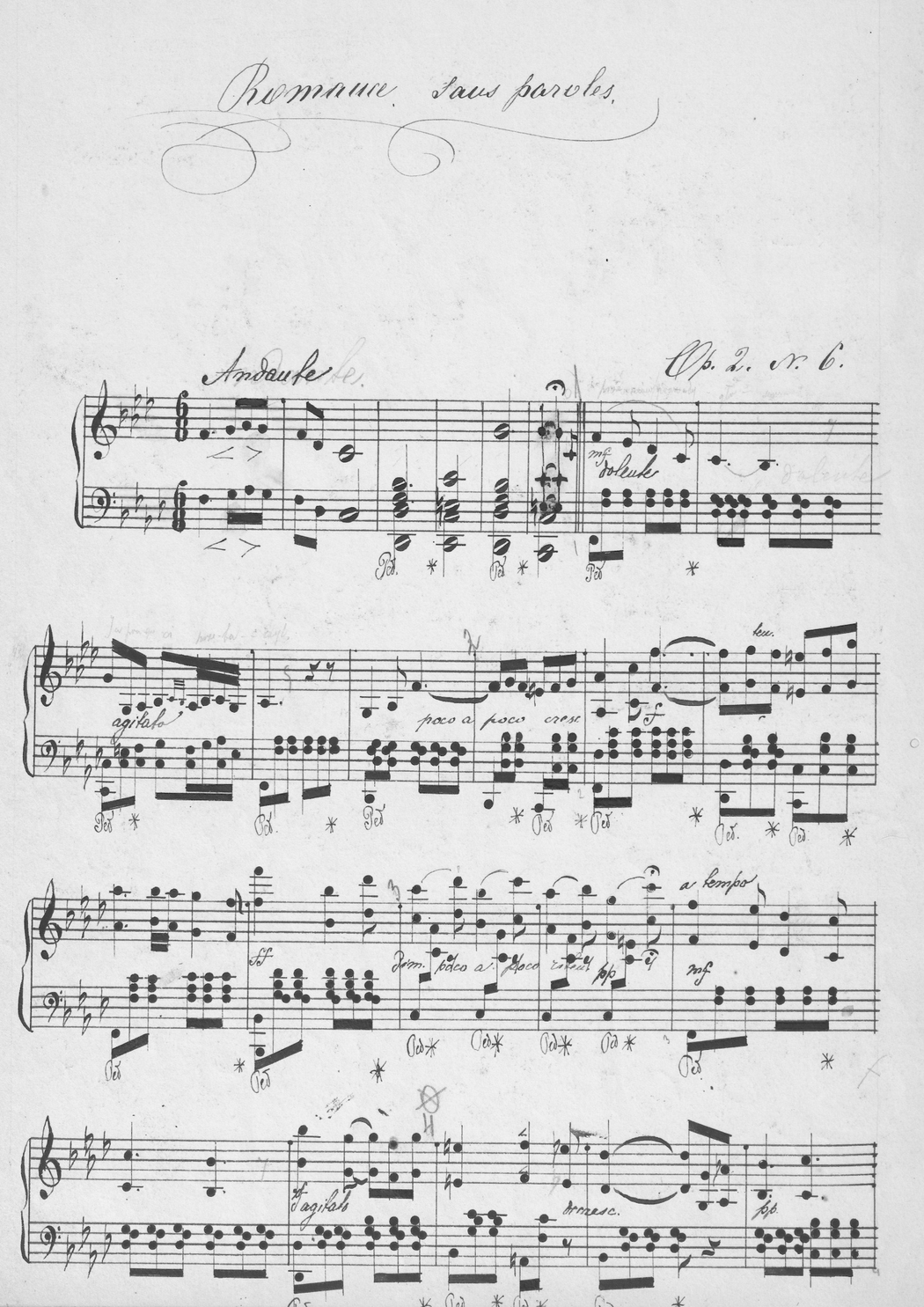}
        \caption{}
    \end{subfigure}\hfill
    \begin{subfigure}[t]{0.495\textwidth}
        \centering
        \includegraphics[width=\linewidth,height=0.36\textheight,keepaspectratio]{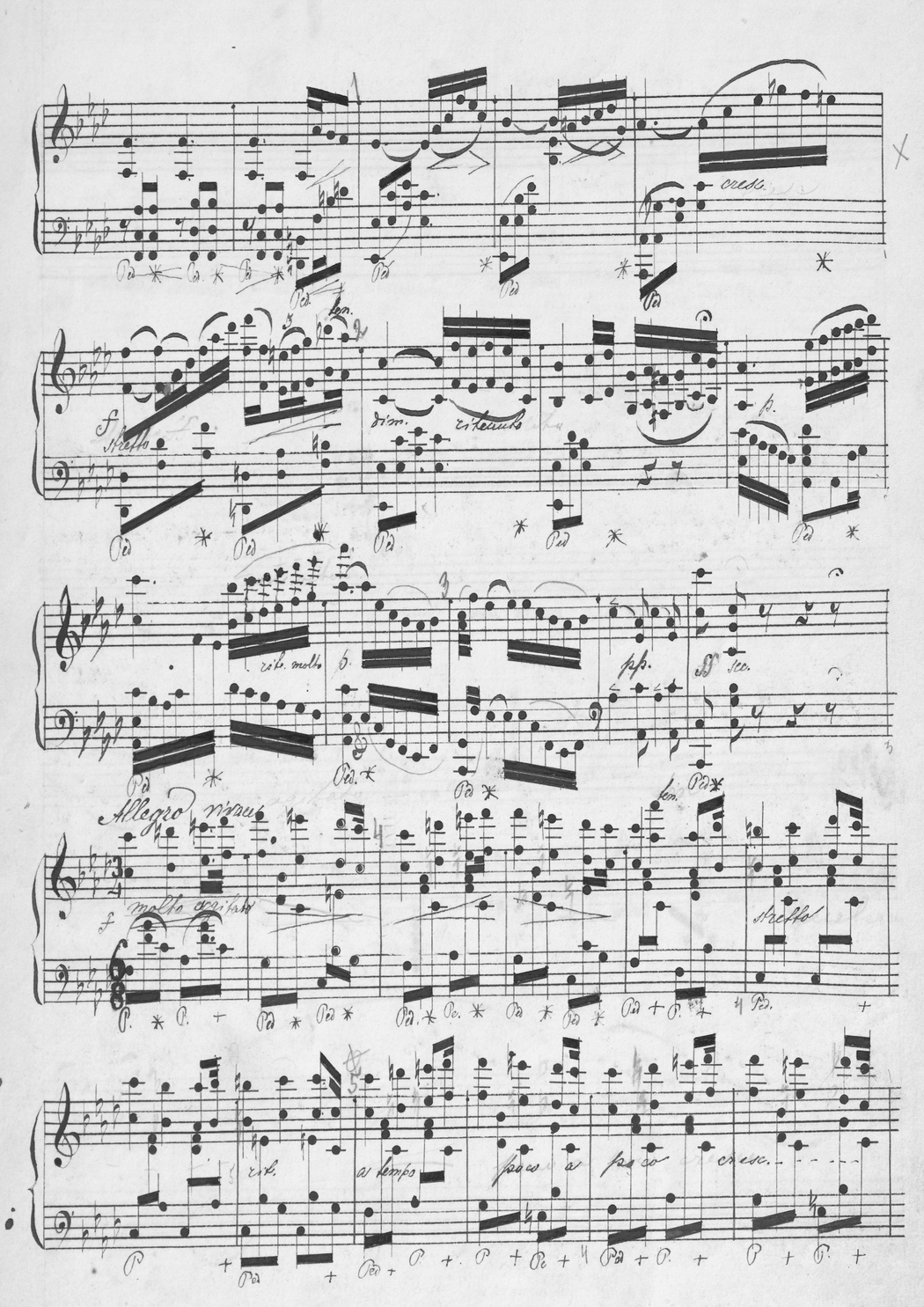}
        \caption{}
    \end{subfigure}

    \vspace{6pt}

    \begin{subfigure}[t]{0.495\textwidth}
        \centering
        \includegraphics[width=\linewidth,height=0.36\textheight,keepaspectratio]{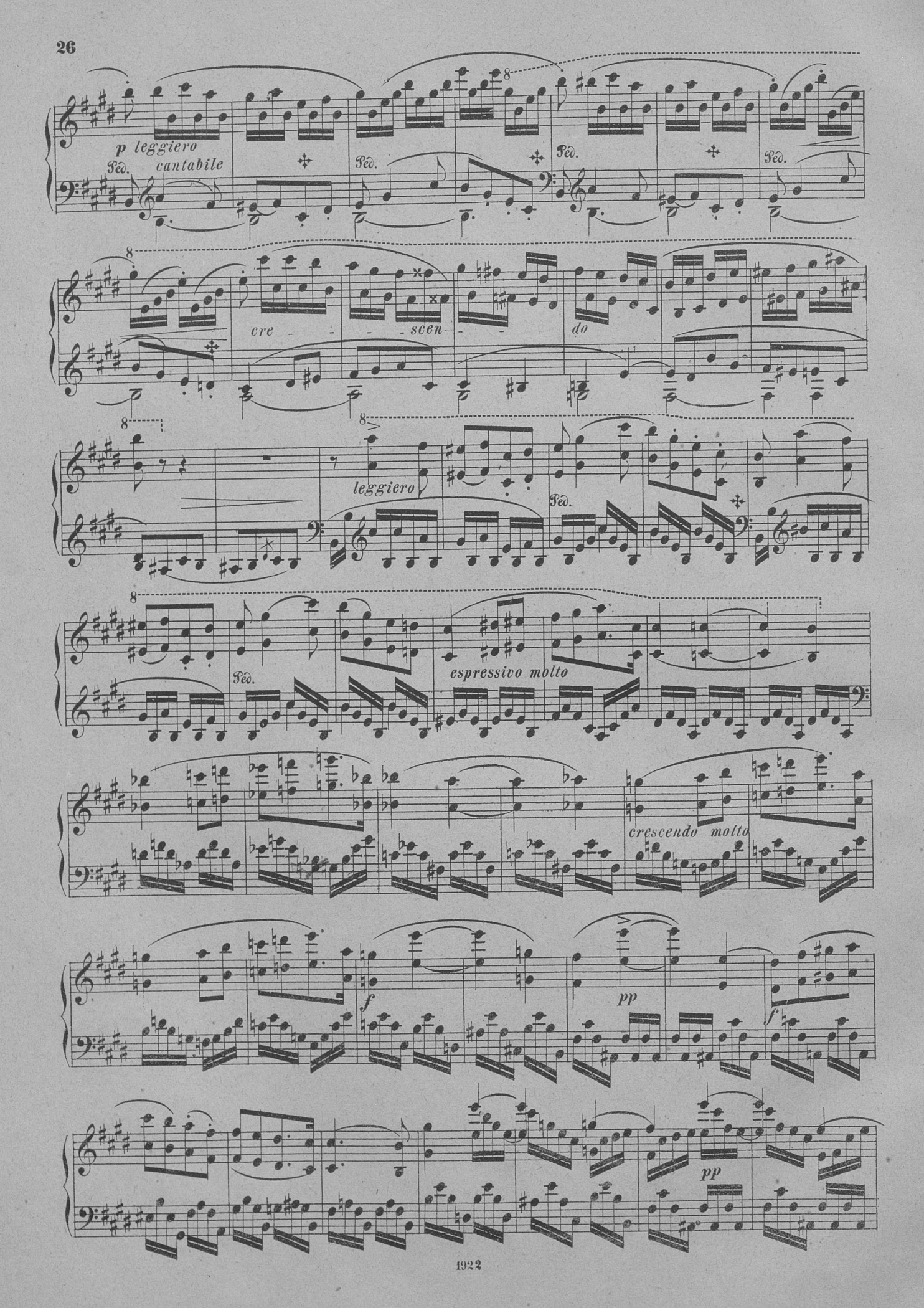}
        \caption{}
    \end{subfigure}\hfill
    \begin{subfigure}[t]{0.495\textwidth}
        \centering
        \includegraphics[width=\linewidth,height=0.36\textheight,keepaspectratio]{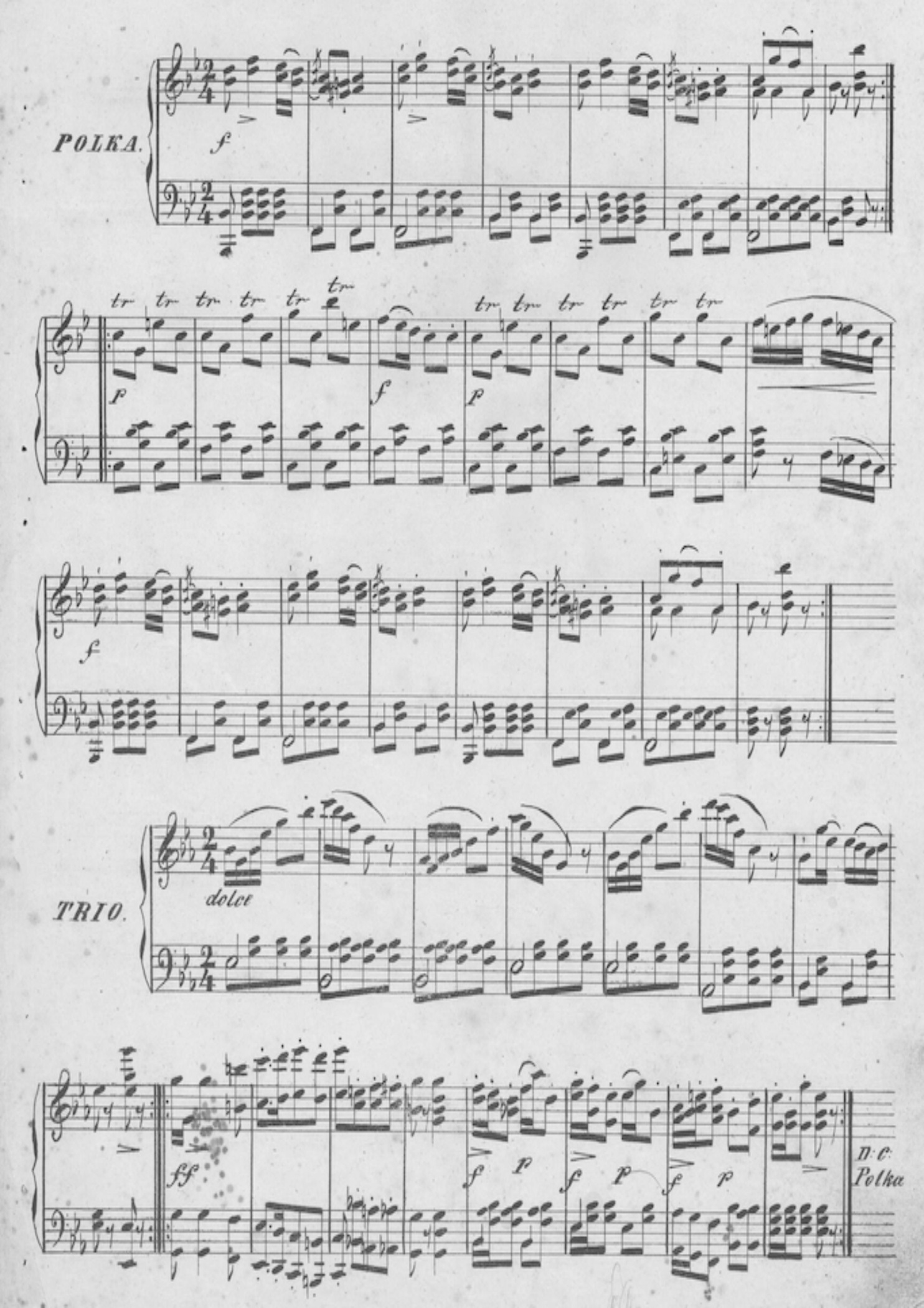}
        \caption{}
    \end{subfigure}

    \caption{Examples from the historical Polish scan benchmark.}
    \label{fig:polish_examples}
\end{figure}
\end{document}

%% file: figures/teaser.tex
\begin{center}
    \captionsetup{type=figure}
    \centering
    \begin{minipage}{0.98\textwidth}
        \begin{minipage}[c]{0.48\linewidth}
            \centering
            \includegraphics[width=\linewidth]{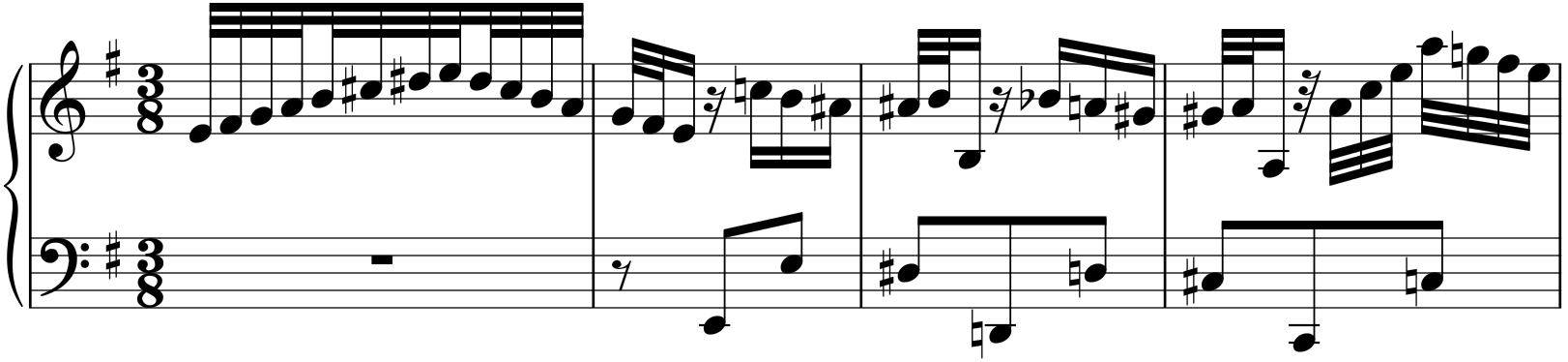}

            \vspace{1pt}
            \begin{tikzpicture}
                \draw[-{Stealth[length=4pt,width=5pt]}, line width=0.45pt, gray!70] (0,0.12) -- (0,-0.12);
            \end{tikzpicture}

            \vspace{2pt}
            \scriptsize
            \setlength{\tabcolsep}{1.6pt}
            \renewcommand{\arraystretch}{0.9}
            \begin{tabular}{@{}l@{\hspace{0.45em}}l@{}}
                \ttfamily \detokenize{**kern}  & \ttfamily \detokenize{**kern}  \\
                \ttfamily \detokenize{*clefF4} & \ttfamily \detokenize{*clefG2} \\
                \ttfamily \detokenize{4.r}     & \ttfamily \detokenize{32eLLL}  \\
                \ttfamily \detokenize{.}       & \ttfamily \detokenize{32f#}    \\
                \ttfamily \detokenize{...}    &
                \ttfamily \detokenize{...}    \\
            \end{tabular}
        \end{minipage}\hfill
        \begin{minipage}[c]{0.48\linewidth}
            \centering
            \begin{tikzpicture}
                \begin{axis}[
                        ybar,
                        width=0.76\linewidth,
                        height=3.2cm,
                        scale only axis,
                        font=\small,
                        ylabel={OMR-NED (\%) $\downarrow$},
                        ylabel style={font=\scriptsize, xshift=5pt},
                        ymin=0,
                        ymax=110,
                        xtick=data,
                        symbolic x coords={Synthetic\\(Verovio), Real Scans\\(Polish)},
                        x tick label style={font=\scriptsize, align=center},
                        legend style={
                                at={(0.5,1.04)},
                                anchor=south,
                                legend columns=3,
                                font=\tiny,
                                draw=none,
                                /tikz/every even column/.append style={column sep=1pt},
                            },
                        enlarge x limits=0.38,
                        grid=major,
                        grid style={gray!20},
                        ytick={0,20,40,60,80,100},
                        nodes near coords,
                        nodes near coords style={font=\tiny, scale=0.75, above},
                    ]
                    \addplot[bar width=13pt, bar shift=-14pt, fill=blue!30, draw=blue!60] coordinates {
                            (Synthetic\\(Verovio),92.23) (Real Scans\\(Polish),80.16)
                        };
                    \addplot[bar width=13pt, bar shift=0pt, fill=orange!40, draw=orange!70] coordinates {
                            (Synthetic\\(Verovio),43.91) (Real Scans\\(Polish),86.73)
                        };
                    \addplot[bar width=13pt, bar shift=14pt, fill=violet!40, draw=violet!70] coordinates {
                            (Synthetic\\(Verovio),18.46) (Real Scans\\(Polish),63.97)
                        };
                    \legend{SMT++, LEGATO, Transcoda (ours)}
                \end{axis}
            \end{tikzpicture}
        \end{minipage}
    \end{minipage}
    \caption{Transcoda maps score images to normalized \texttt{**kern} sequences. Trained entirely on synthetic data, it achieves the lowest error (OMR-NED) among compared systems on both clean synthetic benchmarks and zero-shot transfer to real historical scans.}
    \label{fig:teaser}
\end{center}

%% file: figures/architecture.tex
\begin{center}
    \captionsetup{type=figure}
    \centering
    \definecolor{enccolor}{HTML}{4A90D9}
    \definecolor{pecolor}{HTML}{5BA85B}
    \definecolor{projcolor}{HTML}{888888}
    \definecolor{deccolor}{HTML}{E8913A}
    \definecolor{sattncolor}{HTML}{D46A4A}
    \definecolor{cattncolor}{HTML}{E8913A}
    \definecolor{ffncolor}{HTML}{D4A84A}
    \definecolor{iocolor}{HTML}{8B6DB0}
    \definecolor{gramcolor}{HTML}{C04040}
    \resizebox{0.98\linewidth}{!}{%
        \begin{tikzpicture}[
            block/.style={draw=#1!70, fill=#1!8, rounded corners=4pt, minimum height=1.8cm, minimum width=2.8cm, font=\small, align=center, line width=0.8pt},
            sublayer/.style={draw=#1!70, fill=#1!15, rounded corners=3pt, minimum height=0.75cm, minimum width=3.4cm, font=\footnotesize, align=center, line width=0.6pt},
            dimlab/.style={font=\scriptsize\sffamily, text=black!45},
            arr/.style={-{Stealth[length=3.5pt, width=2.8pt]}, line width=0.9pt, black!50},
            darr/.style={-{Stealth[length=2.5pt, width=2.2pt]}, line width=0.6pt, black!35},
        ]
            \node[block=iocolor, minimum width=2.2cm, minimum height=2.4cm] (input) {\includegraphics[width=1.7cm]{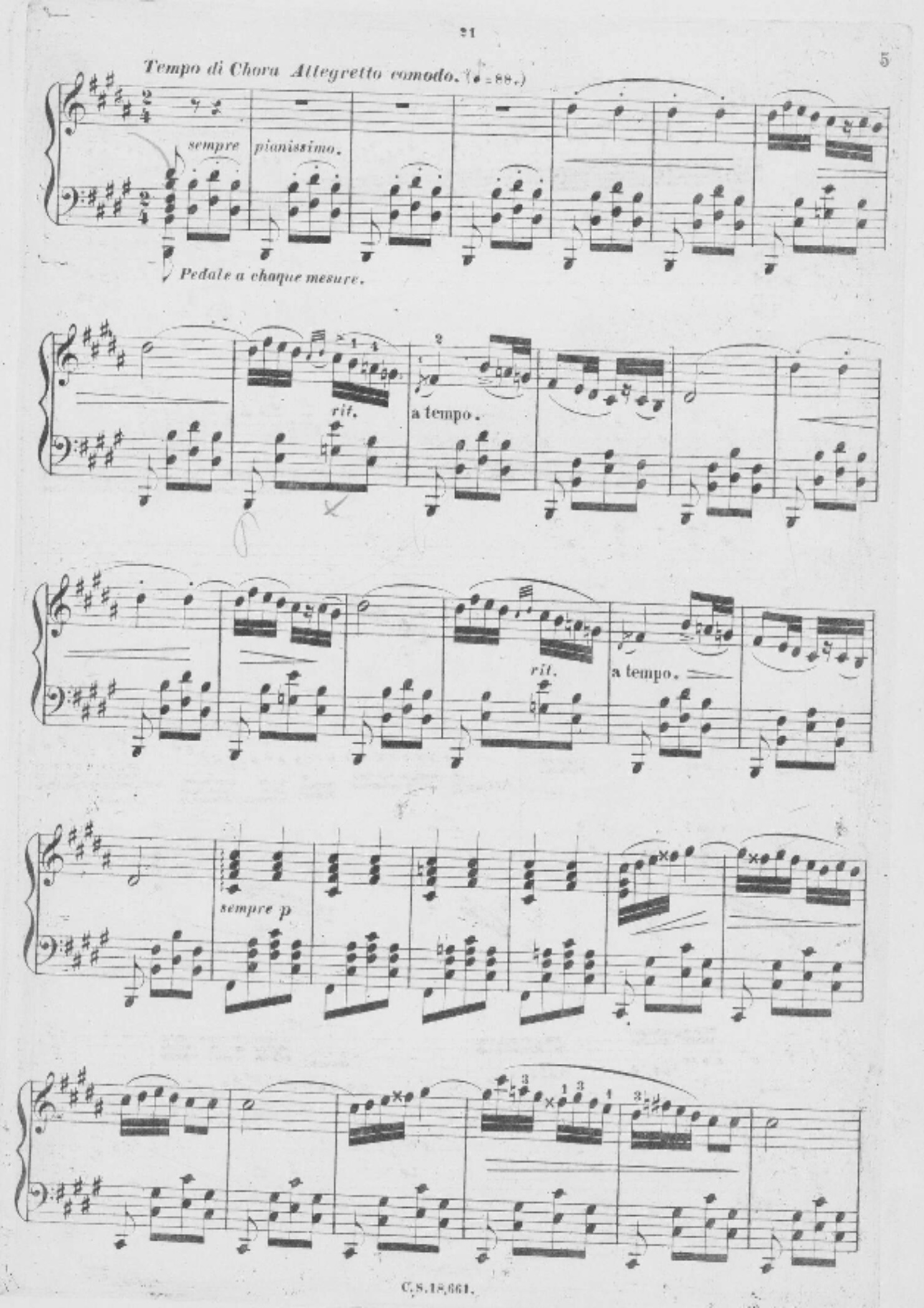}};
            \node[dimlab, below=3pt of input] {$1485{\times}1050$ px (fixed)};
            \node[font=\footnotesize, above=3pt of input] {Score image};

            \node[block=enccolor, right=1.0cm of input, minimum width=3.0cm, minimum height=2.4cm]
            (encoder) {\textbf{ConvNeXt-V2}\\[2pt]{\footnotesize pretrained \& finetuned}};
            \node[dimlab, below=3pt of encoder] {$47{\times}33{\times}768$};
            \node[font=\footnotesize, text=enccolor!80!black, above=3pt of encoder] {Vision encoder};

            \node[block=projcolor, right=0.8cm of encoder, minimum width=2.6cm, minimum height=2.4cm]
            (proj) {\textbf{Projector MLP}\\[2pt]{\footnotesize Lin.\ $\to$ GELU $\to$ Lin.}};
            \node[dimlab, below=3pt of proj] {$1551{\times}512$};

            \node[draw=pecolor!70, fill=pecolor!15, circle, minimum size=0.9cm, font=\normalsize\bfseries, line width=0.8pt, right=1.0cm of proj] (addcircle) {$+$};
            \node[dimlab, below=3pt of addcircle] {$1551{\times}512$};
            \node[block=pecolor, minimum width=2.4cm, minimum height=1.2cm, above=0.8cm of addcircle] (pe) {\textbf{2D Sin.\ PE}};
            \draw[arr, pecolor!60] (pe) -- (addcircle);

            \node[sublayer=sattncolor, right=1.4cm of addcircle, yshift=1.05cm] (selfattn) {\textbf{Self-attention}~~{\scriptsize RoPE}};
            \node[sublayer=cattncolor, below=0.28cm of selfattn] (crossattn) {\textbf{Cross-attention}};
            \node[sublayer=ffncolor, below=0.28cm of crossattn] (ffn) {\textbf{Feed-forward}};

            \coordinate (decoderTop) at ($(selfattn.north)+(0,13pt)$);
            \begin{scope}[on background layer]
                \node[draw=deccolor!40, fill=deccolor!4, rounded corners=5pt, line width=0.9pt, fit=(decoderTop)(selfattn)(crossattn)(ffn), inner xsep=10pt, inner ysep=6pt] (decoder) {};
            \end{scope}
            \draw[decorate, decoration={brace, amplitude=4pt}, line width=0.6pt, black!40]
            ($(decoder.north east)+(3pt,0)$) -- ($(decoder.south east)+(3pt,0)$)
            node[midway, right=6pt, font=\footnotesize] {$\times 8$};
            \draw[darr] (selfattn) -- (crossattn);
            \draw[darr] (crossattn) -- (ffn);
            \node[font=\footnotesize, text=deccolor!80!black, above=6pt of selfattn] (toklabel) {Transformer decoder};
            \draw[arr, iocolor!60] (toklabel) -- (selfattn.north);
            \node[dimlab, below=3pt of decoder] {$d_{\mathrm{model}}{=}512$, $8$ heads, $d_{\mathrm{ff}}{=}1024$};

            \node[draw=gramcolor!70, fill=gramcolor!15, circle, minimum size=0.9cm, font=\normalsize\bfseries, line width=0.8pt, right=1.4cm of decoder.east |- ffn] (maskcircle) {$\otimes$};
            \node[block=gramcolor, minimum width=2.4cm, minimum height=1.2cm, below=0.5cm of maskcircle] (grammar) {\textbf{Constrained}\\\textbf{decoding}};
            \draw[arr, gramcolor!60] (grammar) -- (maskcircle);

            \node[block=iocolor, right=1.3cm of maskcircle, minimum width=2.0cm, minimum height=1.4cm] (output) {\textbf{\texttt{**kern}}\\[2pt]{\footnotesize tokens}};
            \node[dimlab, below=3pt of output] {vocab${=}3000$};

            \draw[arr] (input) -- (encoder);
            \draw[arr] (encoder) -- (proj);
            \draw[arr] (proj) -- (addcircle);
            \draw[arr] (addcircle) -- (crossattn.west);
            \draw[arr] (ffn.east) -- (maskcircle);
            \draw[arr] (maskcircle) -- (output);

            \coordinate (looptop) at ($(toklabel.north)+(0,0.42cm)$);
            \draw[arr, iocolor!50, rounded corners=5pt]
            (output.north) -- node[font=\footnotesize, right=2pt, pos=0.1] {Output} (output.north |- looptop)
            -- node[font=\scriptsize\sffamily, text=iocolor!80!black, above=1pt] {autoregressive} (toklabel.north |- looptop)
            -- (decoder.north);
        \end{tikzpicture}%
    }
    \caption{Transcoda architecture. A ConvNeXt-V2 encoder feeds projected visual features with 2D positional encodings into an 8-layer Transformer decoder. The optional constraint engine masks invalid \texttt{**kern} continuations at inference time.}
    \label{fig:architecture}
\end{center}

%% file: figures/kern_format.tex
\begin{wrapfigure}[23]{r}{0.5\linewidth}
    \vspace{-1.1\baselineskip}
    \centering
    \makebox[\linewidth][c]{%
        \begin{minipage}[t]{0.43\linewidth}
            \raggedleft
            \vspace{0pt}
            \includegraphics[height=7.35cm]{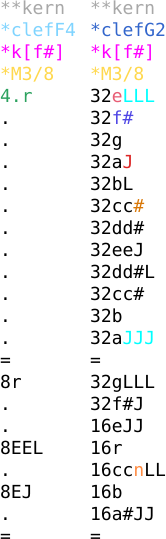}
        \end{minipage}%
        \hspace{0.02\linewidth}%
        \begin{minipage}[t]{0.51\linewidth}
            \raggedright
            \vspace{0pt}
            \rotatebox{-90}{\includegraphics[width=7.35cm]{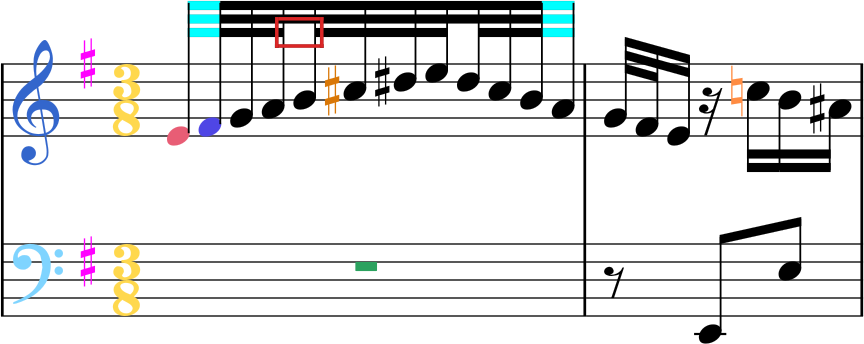}}
        \end{minipage}%
    }
    \caption{The \texttt{**kern} format encodes sheet music as a text grid. Rows represent simultaneous time steps and columns represent parallel voices. Note tokens define pitch and duration, while dot tokens (\texttt{.}) act as placeholders to keep the voices aligned across different rhythms.}
    \label{fig:kern_format}
\end{wrapfigure}

%% file: figures/normalization.tex
\begin{wrapfigure}[15]{r}{0.3\linewidth}
    \vspace{-2.2\baselineskip}
    \centering
    \definecolor{fadegray}{HTML}{8A8A8A}
    \definecolor{canongreen}{HTML}{3C8D4E}
    \resizebox{\linewidth}{!}{%
        \begin{tikzpicture}[
            snippet/.style={draw=black!20, fill=black!2, rounded corners=2pt,
                    inner xsep=4pt, inner ysep=3pt, font=\ttfamily\tiny, align=left},
            canonical/.style={draw=canongreen!60, fill=canongreen!8, rounded corners=2pt,
                    inner xsep=4pt, inner ysep=3pt, font=\ttfamily\tiny\bfseries, align=left},
            arr/.style={-{Latex[length=2mm]}, line width=0.6pt, black!40},
            ]
            \node[inner sep=0pt] (img) {\includegraphics[height=1.3cm]{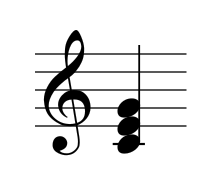}};

            \node[snippet, below left=8mm and 0mm of img, text=fadegray] (v1)
            {\detokenize{4g 4e 4c}};
            \node[snippet, below=8mm of img, text=fadegray] (v2)
            {\detokenize{4e 4c 4g}};
            \node[snippet, below right=8mm and 0mm of img, text=fadegray] (v3)
            {\detokenize{4c 4g 4e}};

            \draw[arr, fadegray] (img.south) -- (v1.north);
            \draw[arr, fadegray] (img.south) -- (v2.north);
            \draw[arr, fadegray] (img.south) -- (v3.north);

            \node[canonical, below=10mm of v2] (canon)
            {\detokenize{4c 4e 4g}};

            \draw[arr, canongreen!70, line width=0.8pt] (v1.south) -- (canon.north west);
            \draw[arr, canongreen!70, line width=0.8pt] (v2.south) -- (canon.north);
            \draw[arr, canongreen!70, line width=0.8pt] (v3.south) -- (canon.north east);
        \end{tikzpicture}%
    }
    \caption{A single visual chord can
        have multiple valid \texttt{**kern} strings. We sort pitches into one normalized sequence to reduce uncertainty.}
    \label{fig:canonicalization}
\end{wrapfigure}

%% file: figures/semantic_augmentation.tex
\begin{figure}[htbp]
    \centering
    \begin{subfigure}[t]{0.485\textwidth}
        \centering
        \includegraphics[width=\linewidth]{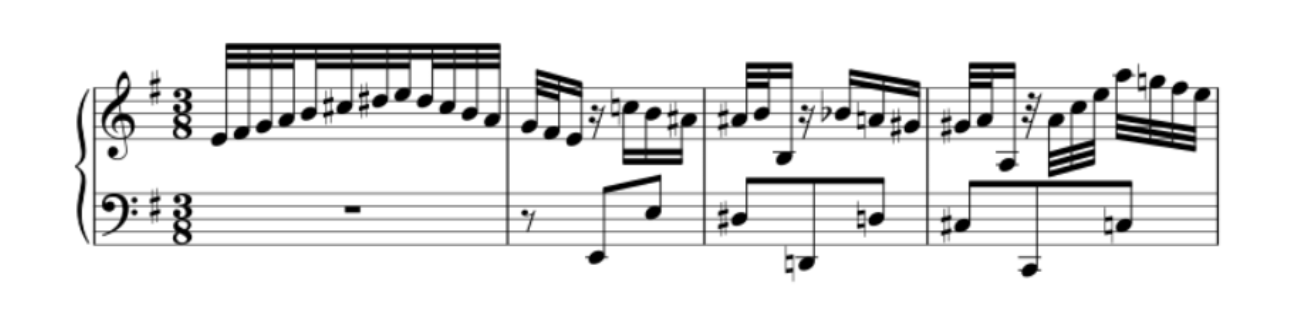}
        \caption{Base render.}
    \end{subfigure}\hfill
    \begin{subfigure}[t]{0.485\textwidth}
        \centering
        \includegraphics[width=\linewidth]{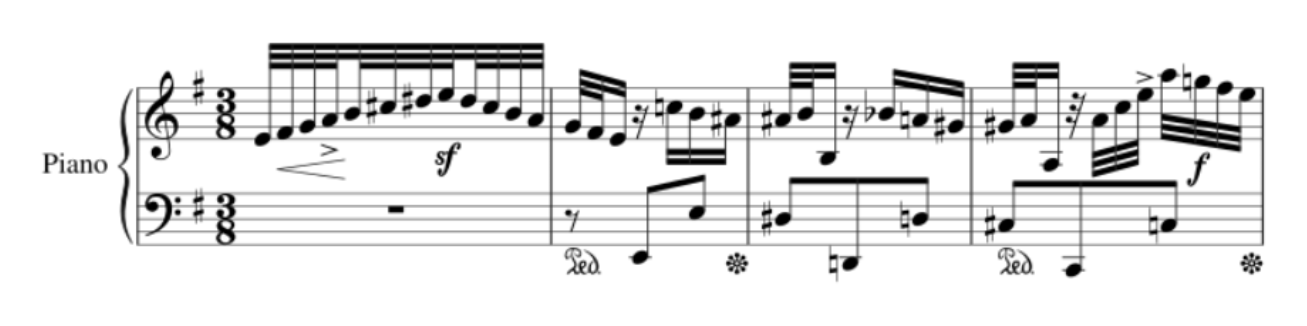}
        \caption{Notation injection.}
    \end{subfigure}

    \vspace{3pt}

    \begin{subfigure}[t]{0.485\textwidth}
        \centering
        \includegraphics[width=\linewidth]{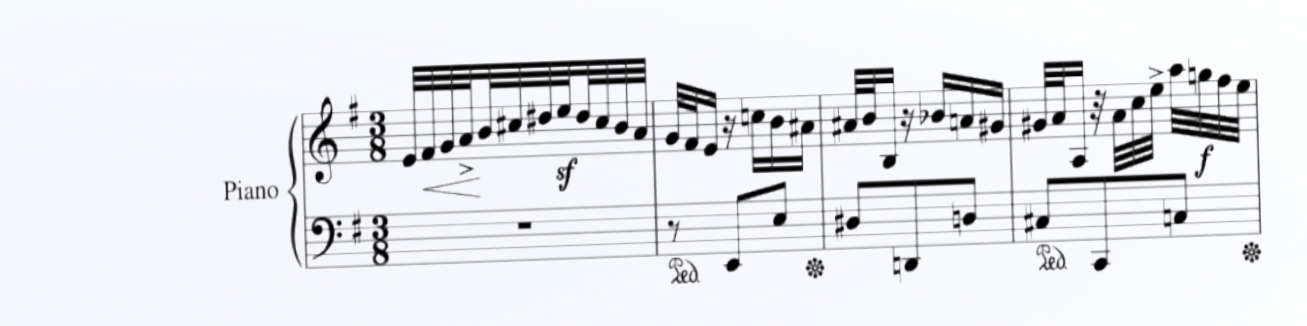}
        \caption{Geometric augmentation.}
    \end{subfigure}\hfill
    \begin{subfigure}[t]{0.485\textwidth}
        \centering
        \includegraphics[width=\linewidth]{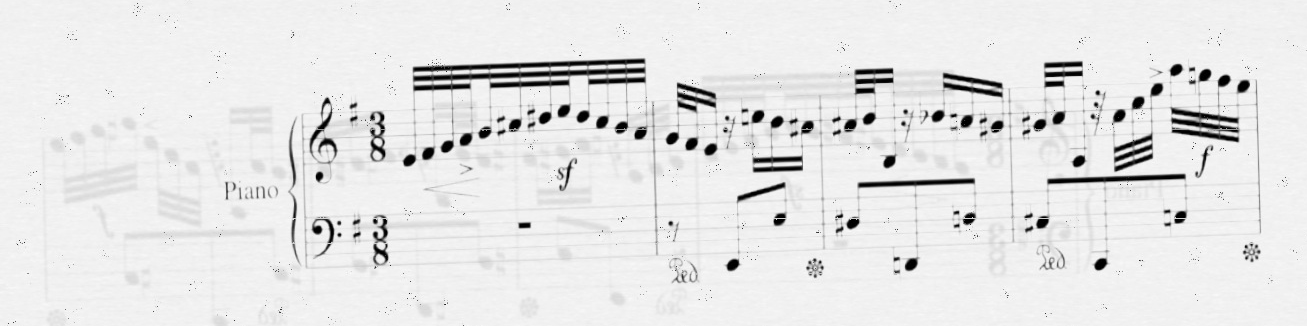}
        \caption{Visual degradation.}
    \end{subfigure}

    \caption{\textbf{Multi-stage data augmentation.} To bridge the sim-to-real gap without altering the \texttt{**kern} target, we apply sequential transformations from stage 4 of our data pipeline.
    } \label{fig:semantic_augmentation}
\end{figure}

%% file: figures/dataset_yield.tex
\begin{wraptable}[18]{r}{0.5\linewidth}
    \vspace{-1.2\baselineskip}
    \centering
    \caption{Dataset yield through textual preprocessing. Normalized files form the source pool for synthetic rendering.}
    \label{tab:data_yield}
    \footnotesize
    \setlength{\tabcolsep}{2.1pt}
    \begin{tabular}{@{}lrrrr@{}}
        \toprule
        Source                                     & Raw     & Conv.   & Filt.   & Norm.   \\
        \specialrule{0.08em}{0.55ex}{0.35ex}
        \multicolumn{5}{@{}l}{\textbf{Training split}}                              \\
        \midrule
        PDMX \cite{long2024pdmx, xu2024generating} & 234,539 & 234,379 & 176,838 & 174,345 \\
        Grandstaff \cite{RiosVila2026}             & 41,598  & 41,598  & 40,859  & 40,859  \\
        MuseTrainer \cite{musetrainer}             & 69      & 69      & 49      & 48      \\
        OpenScore Lieder \cite{OpenScoreLieder}    & 1,389   & 1,389   & 1,017   & 1,008   \\
        OpenScore Quartets \cite{OpenScoreQuartet} & 122     & 121     & 41      & 38      \\
        \midrule
        Total                                      & 277,717 & 277,556 & 218,804 & 216,298 \\
        \specialrule{0.1em}{0.75ex}{0.35ex}
        \multicolumn{5}{@{}l}{\textbf{Validation split}}                            \\
        \midrule
        PDMX \cite{long2024pdmx, xu2024generating} & 6,451   & 6,448   & 2,973   & 2,955   \\
        Grandstaff \cite{RiosVila2026}             & 4,623   & 4,623   & 4,526   & 4,526   \\
        Polish Scores \cite{PRAIGPolishScores}     & 117     & 117     & 104     & 102     \\
        \midrule
        Total                                      & 11,191  & 11,188  & 7,603   & 7,583   \\
        \bottomrule
    \end{tabular}
\end{wraptable}

%% file: figures/ablation_results.tex
\begin{wraptable}[26]{r}{0.5\linewidth}
    \centering
    \caption{Ablation results. The reference row is Transcoda with greedy decoding.}
    \label{tab:ablations}
    \begin{subtable}{\linewidth}
        \centering
        \caption{Synthetic (Verovio)}
        \label{tab:ablations_indomain}
        \setlength{\tabcolsep}{2.2pt}
        \begin{tabular}{@{}lrr@{}}
            \toprule
            Configuration          & CER $\downarrow$ & OMR-NED $\downarrow$ \\
            \midrule
            Transcoda (greedy)     & 4.38             & 18.71                \\
            w/o target norm.       & 27.00            & 82.51                \\
            + constrained decoding & 4.62             & 18.74                \\
            + beam search          & \textbf{2.72}    & \textbf{18.46}       \\
            \bottomrule
        \end{tabular}
    \end{subtable}

    \vspace{0.35em}
    \begin{subtable}{\linewidth}
        \centering
        \caption{Real Scans (Polish)}
        \label{tab:ablations_transfer}
        \setlength{\tabcolsep}{2.2pt}
        \begin{tabular}{@{}lrr@{}}
            \toprule
            Configuration           & CER $\downarrow$ & OMR-NED $\downarrow$ \\
            \midrule
            Transcoda (greedy)      & 33.18            & 65.76                \\
            \midrule
            \multicolumn{3}{@{}l}{\textit{Data engine}}                       \\
            w/o asym. augmentation  & 60.56            & 78.99                \\
            w/o visual degradation  & 41.81            & 76.95                \\
            w/o score concatenation & 46.52            & 80.10
            \\
            \midrule
            \multicolumn{3}{@{}l}{\textit{In-domain pretraining}}             \\
            FCMAE + CNX-V2-Base     & 28.81            & \textbf{60.70}       \\
            \midrule
            \multicolumn{3}{@{}l}{\textit{Inference}}                         \\
            + constrained decoding  & 31.73            & 63.91                \\
            + beam search           & \textbf{27.00}   & 63.97                \\
            \bottomrule
        \end{tabular}
    \end{subtable}
\end{wraptable}

%% file: figures/qualitative.tex
\begin{figure}[!tbp]
    \centering
    \captionsetup{type=figure}
    \centering
    \resizebox{0.82\textwidth}{!}{
    \begin{tikzpicture}[
            row image/.style={
                    inner sep=0pt,
                    outer sep=0pt,
                },
            row label/.style={
                    rotate=90,
                    anchor=center,
                    font=\footnotesize,
                    align=center,
                    minimum width=1.35cm,
                },
        ]
        \def\imgwidth{0.84\textwidth}
        \def\rowsep{1.4mm}
        \def\labeloffset{3.5mm}

        \node[row image] (input) {%
            \includegraphics[width=\imgwidth]{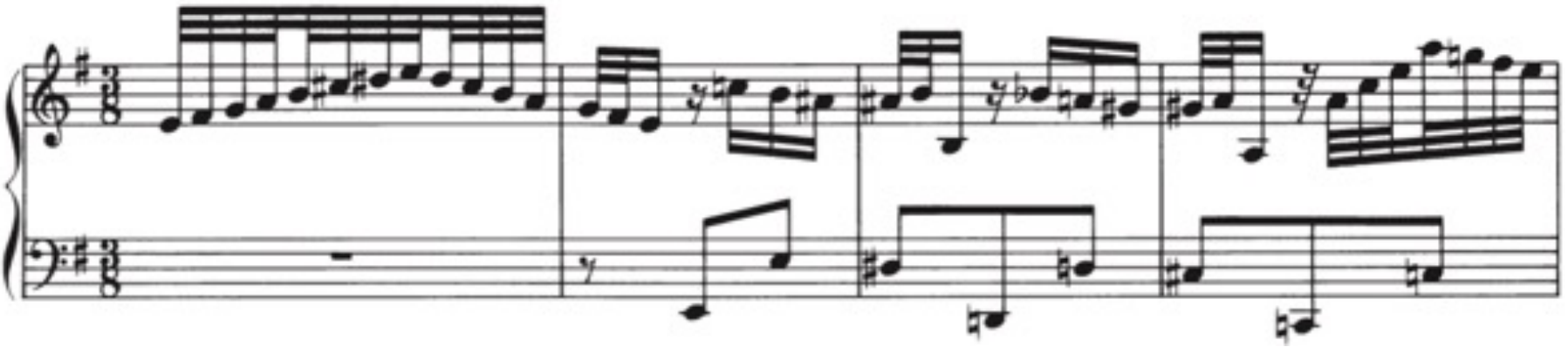}%
        };
        \node[row label] at ([xshift=-\labeloffset]input.west) {\textbf{Input}};

        \node[row image, below=\rowsep of input] (transcoda) {%
            \includegraphics[width=\imgwidth]{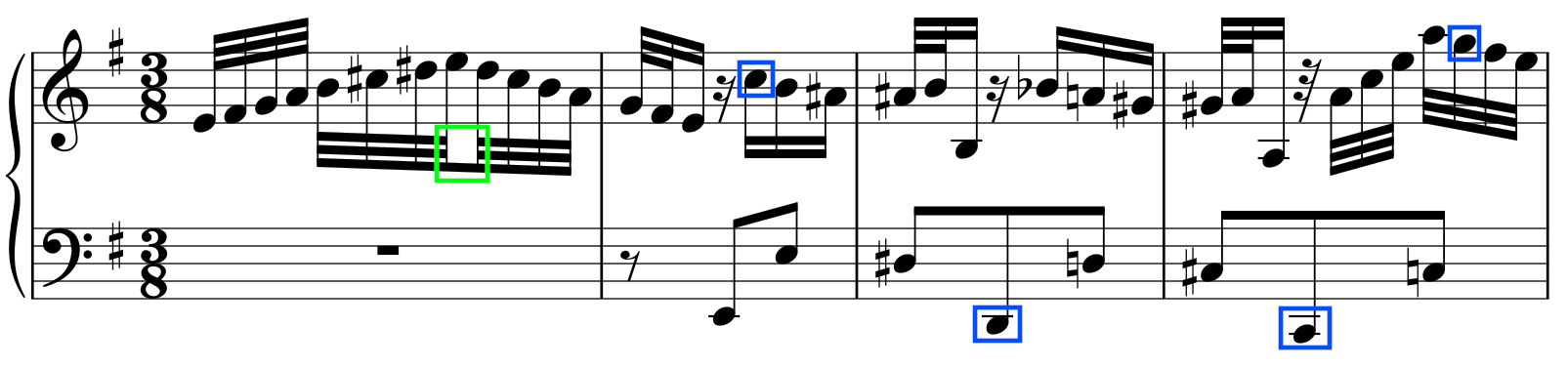}%
        };
        \node[row label] at ([xshift=-\labeloffset]transcoda.west) {%
        \textbf{Transcoda}\\{\scriptsize TEDn: 1.5}%
        };

        \node[row image, below=\rowsep of transcoda] (legato) {%
            \includegraphics[width=\imgwidth]{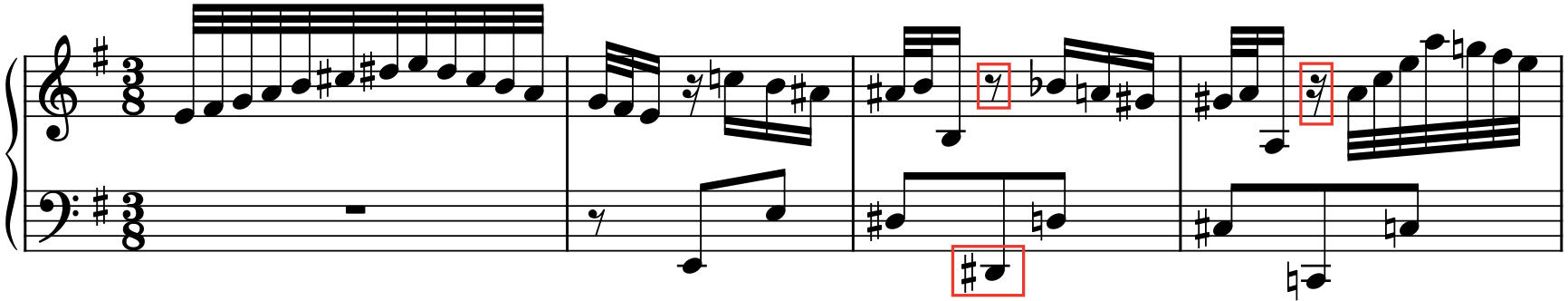}%
        };
        \node[row label] at ([xshift=-\labeloffset]legato.west) {%
        \textbf{Legato}\\{\scriptsize TEDn: 3.12}%
        };

        \node[row image, below=\rowsep of legato] (smtpp) {%
            \includegraphics[width=\imgwidth]{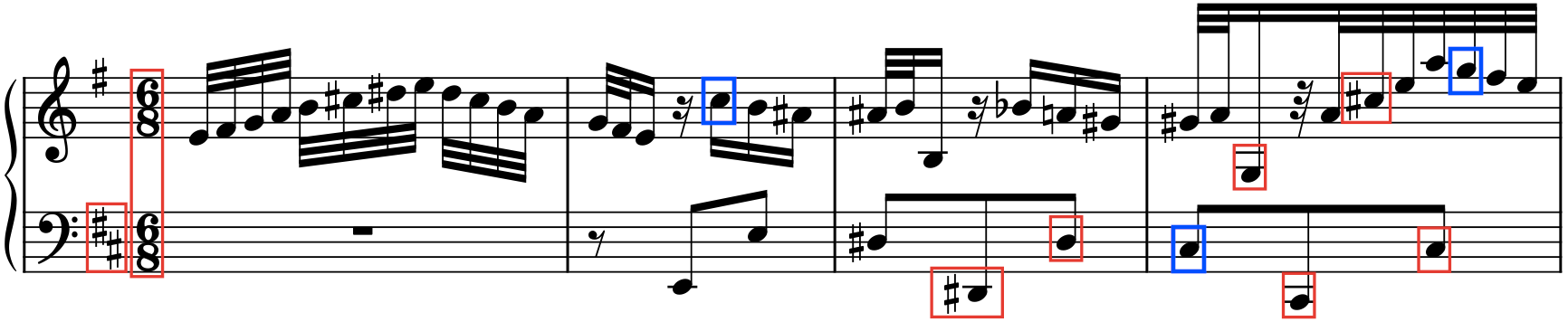}%
        };
        \node[row label] at ([xshift=-\labeloffset]smtpp.west) {%
        \textbf{SMT++}\\{\scriptsize TEDn: 66.04}%
        };
    \end{tikzpicture}
    }
    \caption{Qualitative comparison on a physical scan of Bach's Duetto No.\ 1 in E minor (BWV 802). Transcoda produces the closest rendered transcription among the compared systems on this example, matching the lower TEDn score. Red boxes mark structural or pitch errors, blue boxes mark omitted courtesy accidentals that preserve pitch, and green boxes mark recovered beaming details. Renderings for the Legato and SMT++ baselines are reproduced from \citet{Yang2025Legato}.}
    \label{fig:qualitative}
\end{figure}